\RequirePackage{fix-cm}
\documentclass[smallextended]{svjour3}       
\smartqed  
\usepackage{graphicx}
\usepackage{amsmath}
\usepackage{amsfonts}
\usepackage{amssymb}
\usepackage{booktabs}
\usepackage{array}
\usepackage{url}  
\usepackage{breakcites}
\usepackage{hyperref}
\newcolumntype{P}[1]{>{\centering\arraybackslash}p{#1}}
\begin{document}

\title{Temporal Pattern Attention for Multivariate Time Series Forecasting
}


\author{Shun-Yao Shih*\thanks{* indicates equal contribution. \\ This work was financially supported by the Ministry of Science and Technology of Taiwan.} \and
        Fan-Keng Sun* \and 
        Hung-yi Lee 
}


\institute{Shun-Yao Shih \at
              National Taiwan University \\
              \email{shunyaoshih@gmail.com}           
           \and
           Fan-Keng Sun \at
              National Taiwan University \\
              \email{b03901056@ntu.edu.tw}
            \and
           Hung-yi Lee \at
              National Taiwan University \\
              \email{hungyilee@ntu.edu.tw}
}

\date{Received: date / Accepted: date}

\maketitle

\begin{abstract}
Forecasting of multivariate time series data, for instance the prediction of electricity consumption, solar power production, and polyphonic piano pieces, has numerous valuable applications.
However, complex and non-linear interdependencies between time steps and series complicate this task. 
To obtain accurate prediction, it is crucial to model long-term dependency in time series data, which can be achieved by recurrent neural networks (RNNs) with an attention mechanism.
The typical attention mechanism reviews the information at each previous time step and selects relevant information to help generate the outputs; however, it fails to capture temporal patterns across multiple time steps.
In this paper, we propose using a set of filters to extract time-invariant temporal patterns, similar to transforming time series data into its ``frequency domain''.
Then we propose a novel attention mechanism to select relevant time series, and use its frequency domain information for multivariate forecasting. 
We apply the proposed model on several real-world tasks and achieve state-of-the-art performance in almost all of cases.
Our source code is available at \href{https://github.com/gantheory/TPA-LSTM}{https://github.com/gantheory/TPA-LSTM}.
\end{abstract}

\section{Introduction}\label{sec:introduction}

In everyday life, time series data are everywhere.
We observe evolving variables generated from sensors over discrete time steps and organize them into time series data. 
For example, household electricity consumption, road occupancy rate, currency exchange rate, solar power production, and even music notes can all be seen as time series data.
In most cases, the collected data are often multivariate time series (MTS) data, such as the electricity consumption of multiple clients, which are tracked by the local power company.
There can exist complex dynamic interdependencies between different series that are significant but difficult to capture and analyze.

\begin{figure}[t]
 \centering
 \includegraphics[width=0.75\columnwidth]{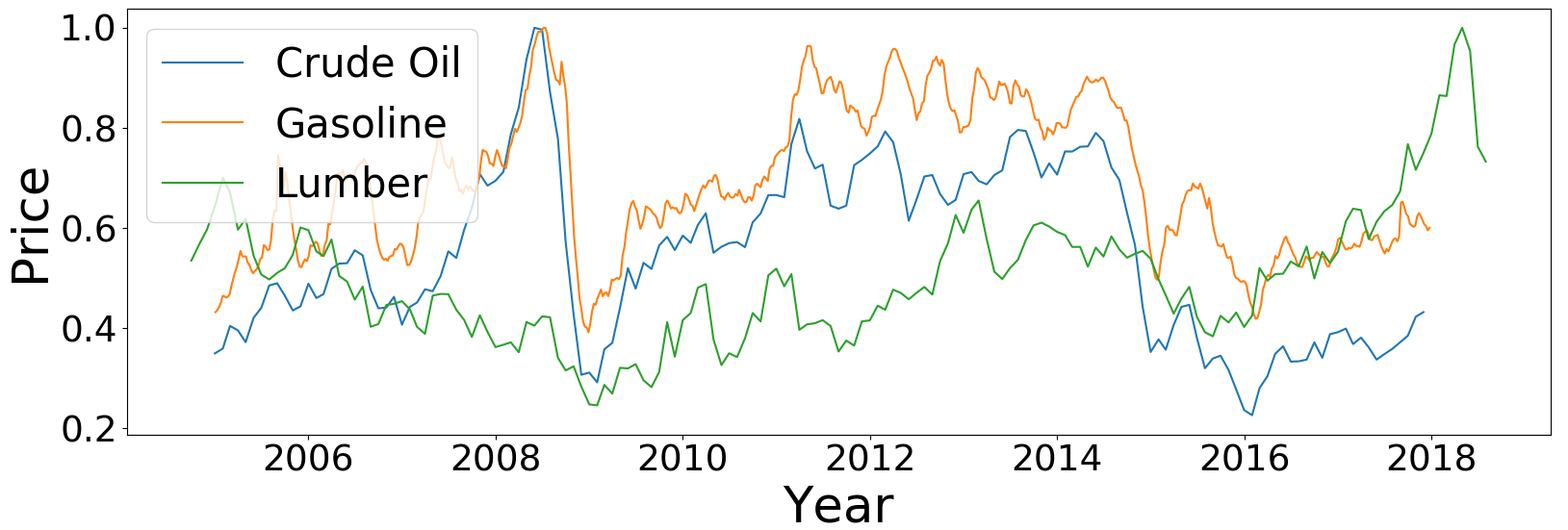}
 \caption{Historical prices of crude oil, gasoline, and lumber. Units are omitted and scales are normalized for simplicity.}
\label{fig:motivation}
\end{figure}

  Analysts   
often seek to forecast the future based on historical data.
The better the interdependencies among different series are modeled, the more accurate the forecasting can be.
For instance, as shown in Figure~\ref{fig:motivation}\footnote{\scriptsize Source: \url{https://www.eia.gov} and \url{https://www.investing.com}}, the price of crude oil heavily influences the price of gasoline, but has a smaller influence on the price of lumber.
Thus, given the realization that gasoline is produced from crude oil and lumber is not, we can use the price of crude oil to predict the price of gasoline.

In machine learning, we want the model to automatically learn such interdependencies from data.
Machine learning has been applied to time series analysis for both classification and forecasting~\cite{forecasting-ANN,forecasting-hybrid-NN,LSTNet,DA-RNN}.
In classification, the machine learns to assign a label to a time series, for instance
evaluating a patient's diagnostic categories by reading values from medical sensors.
In forecasting, the machine predicts future time series based on past observed data.
For example, precipitation in the next days, weeks, or months can be forecast according to historical measurements.
The further ahead we attempt to forecast, the harder it is.

When it comes to MTS forecasting using deep learning, recurrent neural networks (RNNs)~\cite{RNN_0,RNN_1,RNN_2} are often used.
However, one disadvantage in using RNNs in time series analysis is their weakness on managing long-term dependencies, for instance yearly patterns in a daily recorded sequence~\cite{properties-S2S}.
The attention mechanism~\cite{luong,bahdanau}, originally utilized in encoder-decoder~\cite{S2S} networks, somewhat alleviates this problem, and thus boosts the effectiveness of RNN~\cite{LSTNet}. 

In this paper, we propose the temporal pattern attention, a new attention mechanism for MTS forecasting, where we use the term ``temporal pattern'' to refer to any time-invariant pattern across multiple time steps. 
The typical attention mechanism identifies the time steps relevant to the prediction, and extracts the information from these time steps, which poses obvious limitations for MTS prediction.
Consider the example in Figure~\ref{fig:motivation}.
To predict the value of gasoline, the machine must learn to focus on ``crude oil'' and ignore ``lumber''.
In our temporal pattern attention, instead of selecting the relevant time steps as in the typical attention mechanism, the machine learns to select the relevant time series.

In addition, time series data often entails noticeable periodic temporal patterns, which are critical for prediction.
However, the periodic patterns  spanning multiple time steps  are difficult for the typical attention mechanism to identify, as it usually focuses only on a few time steps. 
In temporal pattern attention, we introduce a convolutional neural network (CNN)~\cite{CNN_0,CNN_1} to extract temporal pattern information from each individual variable.

The main contributions of this paper are summarized as follows:
\begin{itemize}
\item We introduce a new attention concept in which we select the relevant variables as opposed to the relevant time steps. The method is simple and general to apply on RNN.
\item We use toy examples to verify that our attention mechanism enables the model to extract temporal patterns and focus on different time steps for different time series.
\item Attested by experimental results on real-world data ranging from periodic and partially linear to non-periodic and non-linear tasks, we show that the proposed attention mechanism achieves state-of-the-art results across multiple datasets.
\item The learned CNN filters in our attention mechanism demonstrate interesting and interpretable behavior.
\end{itemize}

The remainder of this paper is organized as follows.
In Section~\ref{sec:related_work} we review related work and in Section~\ref{sec:preliminaries} we describe background knowledge.
Then, in Section~\ref{sec:temporal_pattern_attention} we describe the proposed attention mechanism.
Next, we present and analyze our attention mechanism on toy examples in Section~\ref{sec:toy_example}, and on MTS and polyphonic music dataset in Section~\ref{sec:experiments_and_analysis}.
Finally, we conclude in Section~\ref{sec:conclusions}.

\section{Related Work}\label{sec:related_work}

The most well-known model for linear univariate time series forecasting is the autoregressive integrated moving average (ARIMA)~\cite{autoregression_0}, which encompasses other autoregressive time series models, including autoregression (AR), moving average (MA), and autoregressive moving average (ARMA). 
Additionally, linear support vector regression (SVR)~\cite{SVR_0,SVR_1} treats the forecasting problem as a typical regression problem with time-varying parameters.
However, these models are mostly limited to linear univariate time series and do not scale well to MTS.
To forecast MTS data, vector autoregression (VAR), a generalization of AR-based models, was proposed.
VAR is probably the most well-known model in MTS forecasting.
Nevertheless, neither AR-based nor VAR-based models capture non-linearity.
For that reason, substantial effort has been put into non-linear models for time series forecasting based on kernel methods~\cite{kernel-method}, ensembles~\cite{ensemble-method}, Gaussian processes~\cite{Gaussian-Process_0} or regime switching~\cite{SETAR}.
Still, these approaches apply predetermined non-linearities and may fail to recognize different forms of non-linearity for different MTS.

Recently, deep neural networks have received a great amount of attention due to their ability to capture non-linear interdependencies.
Long short-term memory (LSTM)~\cite{lstm}, a variant of recurrent neural network, has shown promising results in several NLP tasks and has also been employed for MTS forecasting.
Work in this area began with using naive RNN~\cite{naive-RNN}, improved with hybrid models that combined ARIMA and multilayer perceptrons~\cite{forecasting-ANN,forecasting-hybrid-NN,hybrid}, and then most recently progressed to dynamic Boltzmann machines with RNN~\cite{DyBM}.
Although these models can be applied to MTS, they mainly target univariate or bivariate time series.

\begin{figure*}[t]
    \centering
    \includegraphics[width=\textwidth]{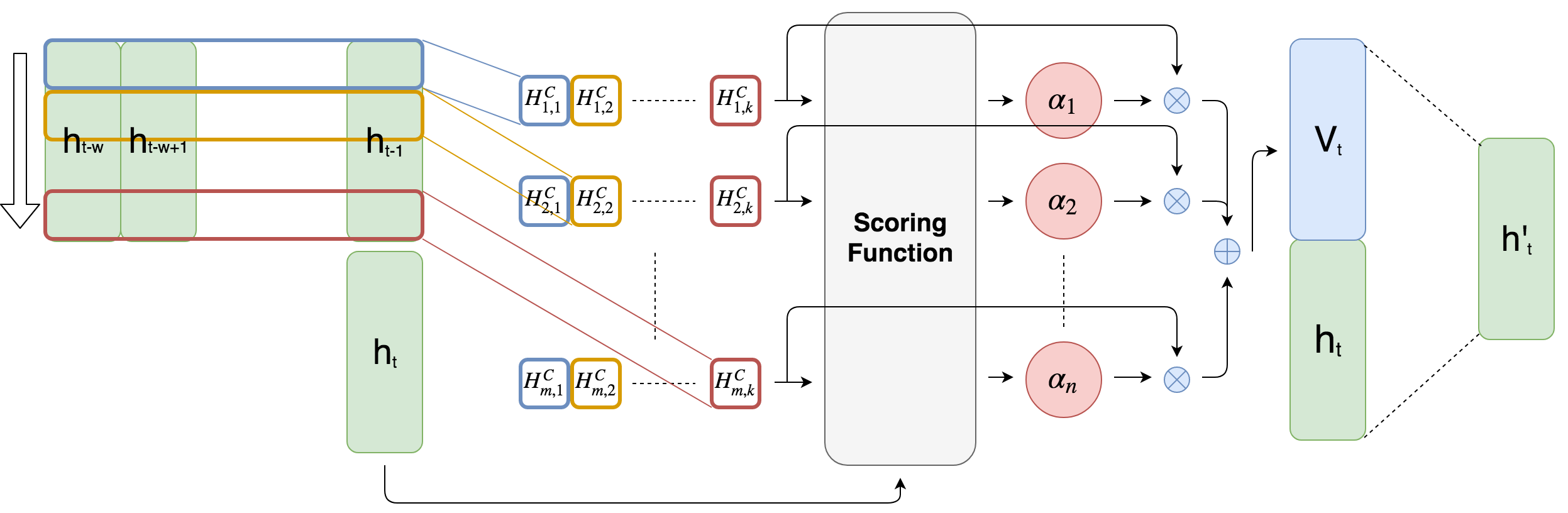}
    \caption{Proposed attention mechanism. $h_t$ represents the hidden state of the RNN at time step $t$. There are $k$ 1-D CNN filters with length $w$, shown as different colors of rectangles. Then, each filter convolves over $m$ features of hidden states and produces a matrix $H^C$ with $m$ rows and $k$ columns. Next, the scoring function calculates a weight for each row of $H^C$ by comparing with the current hidden state $h_t$. Last but not least, the weights are normalized and the rows of $H^C$ is weighted summed by their corresponding weights to generate $V_t$. Finally, we concatenate $V_t, h_t$ and perform matrix multiplication to generate $h^\prime_t$, which is used to create the final forecast value.}
    \label{fig:attention_structure}
\end{figure*}

To the best of our knowledge, the long- and short-term time-series network (LSTNet)~\cite{LSTNet} is the first model designed specifically for MTS forecasting with up to hundreds of time series.
LSTNet uses CNNs to capture short-term patterns, and LSTM or GRU for memorizing relatively long-term patterns.
In practice, however, LSTM and GRU cannot memorize very long-term interdependencies due to training instability and the gradient vanishing problem.
To address this, LSTNet adds either a recurrent-skip layer or a typical attention mechanism.
Also part of the overall model is traditional autoregression, which helps to mitigate the scale insensitivity of neural networks.
Nonetheless, LSTNet has three major shortcomings when compared to our proposed attention mechanism:
(1) the skip length of the recurrent-skip layer must be manually tuned in order to match the period of the data, whereas our proposed approach learns the periodic patterns by itself;
(2) the LSTNet-Skip model is specifically designed for MTS data with periodic patterns, whereas our proposed model, as shown in our experiments, is simple and adaptable to various datasets, even non-periodic ones; and
(3) the attention layer in LSTNet-Attn model selects a relevant hidden state as in typical attention mechanism, whereas our proposed attention mechanism selects relevant time series which is a more suitable mechanism for MTS data.

\section{Preliminaries}\label{sec:preliminaries}

In this section, we briefly describe two essential modules related to our proposed model: the RNN module, and the typical attention mechanism.

\subsection{Recurrent Neural Networks}
Given a sequence of information $\{x_1, x_2, \ldots, x_t\}$, where $x_i \in \mathbb{R}^n$, an RNN generally defines a recurrent function, $F$, and calculates $h_t \in \mathbb{R}^m$ for each time step, $t$, as
\begin{equation}
  h_t = F(h_{t - 1}, x_t)
\end{equation}
where the implementation of 
  function $F$   
depends on what kind of RNN cell is used.

Long short-term memory (LSTM)~\cite{lstm} cells are widely used, which have a slightly different recurrent function:
\begin{equation}
  h_t, c_t = F(h_{t - 1}, c_{t - 1}, x_t), 
\end{equation}
which is defined by the following equations:
\begin{align}
  i_t   &= \textit{sigmoid}(W_{x_i} x_t + W_{h_i} h_{t - 1}) \label{eq:lstm1}\\
  f_t   &= \textit{sigmoid}(W_{x_f} x_t + W_{h_f} h_{t - 1}) \label{eq:lstm2}\\
  o_t   &= \textit{sigmoid}(W_{x_o} x_t + W_{h_o} h_{t - 1}) \label{eq:lstm3}\\
  c_t &= f_t \odot c_{t - 1} + i_t \odot \textit{tanh}(W_{x_g} x_t + W_{h_g} h_{t - 1}) \label{eq:lstm4} \\
  h_t &= o_t \odot \textit{tanh}(c_t)
  \label{eq:lstm5}
\end{align} 
where
$i_t$, $f_t$, $o_t$, and $c_t \in \mathbb{R}^m$,
$W_{x_i}$, $W_{x_f}$, $W_{x_o}$ and $W_{x_g} \in \mathbb{R}^{m \times n}$,
$W_{h_i}$, $W_{h_f}$, $W_{h_o}$ and $W_{h_g} \in \mathbb{R}^{m \times m}$,
and $\odot$ denotes element-wise multiplication.

\subsection{Typical Attention Mechanism}

In the typical attention mechanism~\cite{luong,bahdanau} in an RNN, given the previous states $H = \{h_1, h_2, \ldots, h_{t - 1}\}$, a context vector $v_t$ is extracted from the previous states.
$v_t$ is a weighted sum of each column $h_i$ in $H$, which represents 
  the information relevant to the current time step.
$v_t$ is further integrated with the present state $h_t$ to yield the prediction.

Assume a scoring function $f: \mathbb{R}^m \times \mathbb{R}^m \mapsto \mathbb{R}$ which computes the relevance between its input vectors.
Formally, we have the following formula to calculate the context vector $v_t$:
\begin{equation}
  \alpha_i = \frac{\exp(f(h_i, h_t))}{\sum_{j=1}^{t - 1} \exp(f(h_j, h_t))}
\end{equation}
\begin{equation}
  v_t = \sum_{i=1}^{t - 1} \alpha_i h_i .
\end{equation}

\section{Temporal Pattern Attention}\label{sec:temporal_pattern_attention}

While previous work focuses mainly on changing the network architecture of the attention-based models via different settings to improve performance on various tasks, we believe there is a critical defect in applying typical attention mechanisms on RNN for MTS forecasting. 
The typical attention mechanism selects information relevant to the current time step, and the context vector $v_t$ is the weighted sum of the column vectors of previous RNN hidden states, $H = \{h_1, h_2, \ldots, h_{t - 1}\}$. 
This design lends itself to tasks in which each time step contains a single piece of information, for example, an NLP task in which each time step corresponds to a single word.
If there are multiple variables in each time step, it fails to ignore 
  variables which are noisy in terms of forecasting utility.
Moreover, since the typical attention mechanism averages the information across multiple time steps, it fails to detect temporal patterns useful for forecasting.

The overview of the proposed model is shown in Figure~\ref{fig:attention_structure}.
In the proposed approach, given previous RNN hidden states $H \in \mathbb{R}^{m \times (t - 1)}$, the proposed attention mechanism basically attends to its row vectors.
The attention weights on rows select those variables that are helpful for forecasting. 
Since the context vector $v_t$ is now the weighted sum of the row vectors containing the information across multiple time steps, it captures temporal information.

\subsection{Problem Formulation} \label{subsec:problem_formulation}

In MTS forecasting, given an MTS, $X = \{x_1, x_2, \ldots, x_{t - 1}\}$, where $x_i \in \mathbb{R}^n$ represents the observed value at time $i$, the task is to predict the value of $x_{t - 1 + \Delta}$, where $\Delta$ is a fixed horizon with respect to different tasks.
We denote the corresponding prediction as $y_{t - 1 + \Delta}$, and the ground-truth value as $\hat{y}_{t - 1 + \Delta} = x_{t - 1 + \Delta}$.
Moreover, for every task, we use only $\{x_{t - w}, x_{t - w + 1}, \ldots, x_{t - 1}\}$ to predict $x_{t - 1 + \Delta}$, where $w$ is the window size.
This is a common practice~\cite{LSTNet,DA-RNN}, because the assumption is that there is no useful information before the window and the input is thus fixed.

\subsection{Temporal Pattern Detection using CNN}

CNN's success lies in no small part to its ability to capture various important signal patterns; as such we use a CNN to enhance the learning ability of the model by applying CNN filters on the row vectors of $H$.
Specifically, we have $k$ filters $C_i \in \mathbb{R}^{1 \times T}$, where $T$ is the maximum length we are paying attention to.
If unspecified, we assume $T = w$.
Convolutional operations yield $H^C \in \mathbb{R}^{n \times k}$ where $H^C_{i, j}$ represents the convolutional value of the $i$-th row vector and the $j$-th filter.
Formally, this operation is given by
\begin{equation}
    H^C_{i, j} =  \sum_{l = 1}^{w} H_{i, (t - w - 1 + l)} \times C_{j, T - w + l}.
\end{equation}

\subsection{Proposed Attention Mechanism}\label{ssec:proposed_attention}

We calculate $v_t$ as a weighted sum of row vectors of $H^C$. 
Defined below is the scoring function $f: \mathbb{R}^k \times \mathbb{R}^m \mapsto \mathbb{R}$ to evaluate relevance:
\begin{equation}
    f(H^C_i, h_t) = (H^C_i)^\top W_a h_t,
    \label{eq:scoring}
\end{equation}
where $H^C_i$ is the $i$-th row of $H^C$, and $W_a \in \mathbb{R}^{k \times m}$.
The attention weight $\alpha_i$ is obtained as
\begin{equation}
    \alpha_i = \textit{sigmoid}( f(H^C_i, h_t) ).
\end{equation}
Note that we use the sigmoid activation function instead of softmax, as we expect more than one variable to be useful for forecasting.

Completing the process, the row vectors of $H^C$ are weighted by $\alpha_i$ to obtain the context vector $v_t  \in \mathbb{R}^k $,  
\begin{equation}
  v_t = \sum_{i=1}^{n} \alpha_i H^C_i.
\end{equation}
Then we integrate $v_t$ and $h_t$ to yield the final prediction
\begin{equation}
  h'_t = W_{h} h_t + W_{v} v_t,
\end{equation}
\begin{equation}
  y_{t - 1 + \Delta} = W_{h'} h'_t,
\end{equation}
where $h_t$, $h'_t \in \mathbb{R}^m$, $W_{h} \in \mathbb{R}^{m \times m}$, $W_{v} \in \mathbb{R}^{m \times k}$, and $W_{h'} \in \mathbb{R}^{n \times m}$ and $y_{t - 1 + \Delta} \in \mathbb{R}^n$.

\begin{figure}[t]
  \centering
  \includegraphics[width=0.92\columnwidth]{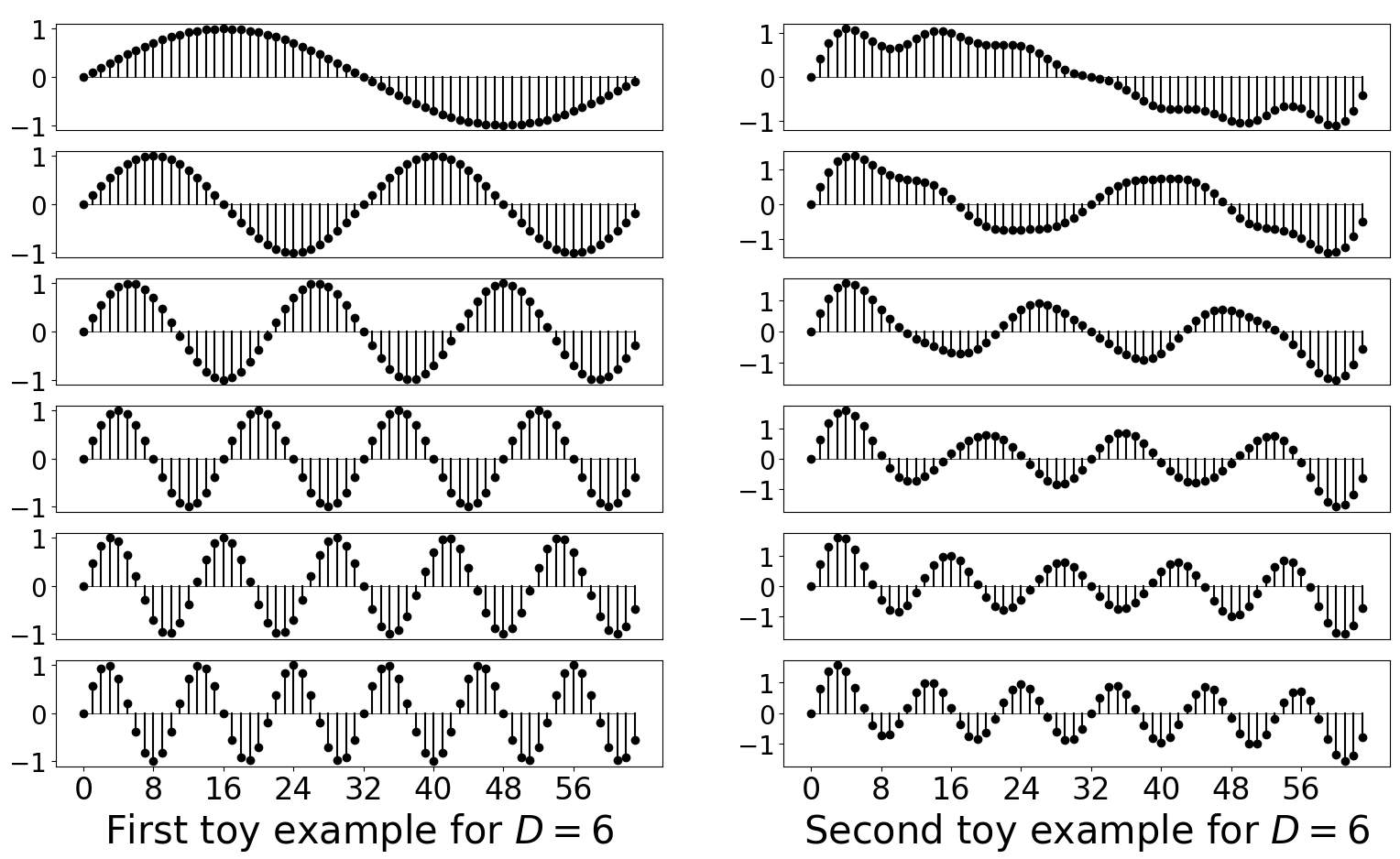}
  \caption{Visualization of the first type of toy examples without interdependencies (left) and the second type of toy examples with interdependencies (right) for $D = 6$, which means that there are 6 time series in each example.}
  \label{fig:vis_toy}
\end{figure}

\begin{figure}[t]
  \centering
  \includegraphics[width=0.83\columnwidth]{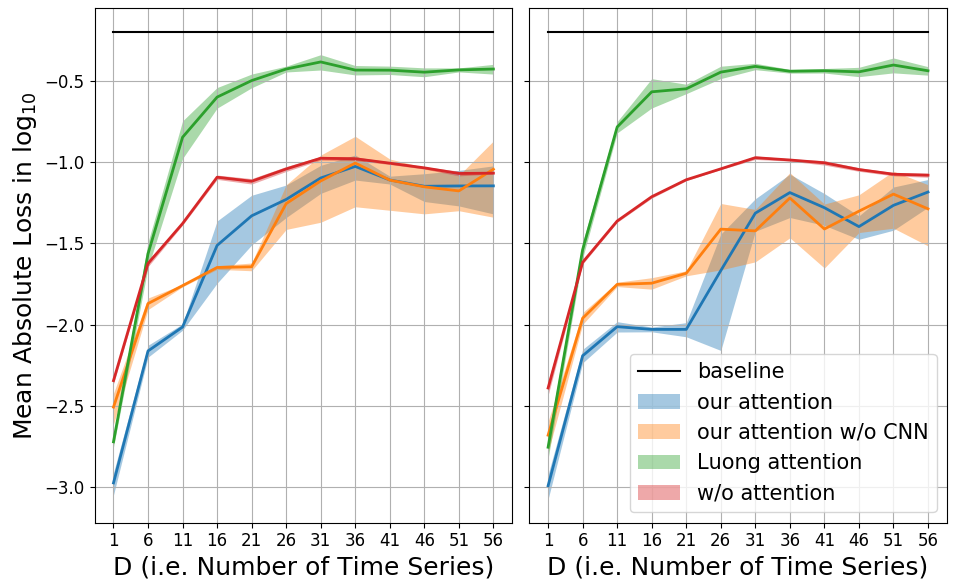}
  \caption{Mean absolute loss and the range of standard deviation in $\log_{10}$ of the first type of toy examples without interdependencies (left) and the second type of toy examples with interdependencies (right), both in ten runs. The baseline indicates the loss if all predicted values are zero.}
  \label{fig:toy}
\end{figure}

\section{Analysis of Proposed Attention on Toy Examples} \label{sec:toy_example}
In order to elaborate the failure of traditional attention mechanisms and the influence of interdependencies, we study the performance of different attention mechanisms on two types of artificially constructed toy examples.

In the first type of toy examples, the $t$-th time step of the $i$-th time series is defined as $sin(\frac{2 \pi i t}{64})$, that is, each time series is a sine wave with different periods.
Notice that any two time series are mutually independent in the first type, so there are no interdependency.

The second type of toy examples adds interdependencies to the first type by mixing time series, and thus the $t$-th time step of the $i$-th time series is formulated as:
\begin{equation}
  sin(\frac{2 \pi i t}{64}) + \frac{1}{D-1} \sum_{j = 1, j \neq i}^D sin(\frac{2 \pi j t}{64}),
  \label{eq:mts}
\end{equation}
where $D$ is the number of time series.
Both types of toy examples are visualized in Fig.~\ref{fig:vis_toy} for $D = 6$.

All models in the following analyses are trained with window size $w = 64$, horizon $\Delta = 1$, and similar amount of parameters.
In this setup, each of our toy examples consists of 64 samples.
Each time series in the first sample comprises values of Eq.~\ref{eq:mts} from t = 0 to 63, and
we can shift one time step to get the second sample with values from t = 1 to 64.
For the last sample, we use values from t = 63 to 126 as the input series correspondingly.
Note that values from t = 64 to 127 are equal to those from t = 0 to 63.
We trained the models for 200 epochs on two types of toy examples for $D = \{1, 6, 11, \dots, 56\}$ and record mean absolute loss in training.
There is no validation and testing data because the intent of this section is to demonstrate the greater capability of our attention over typical attention to fit MTS data not the generalizability of our attention.
The results are shown in Fig.~\ref{fig:toy}.

\subsection{Failure of traditional attention mechanisms}
Intuitively, for the first toy example, the model can accurately predict the next value by memorizing the value that appears exactly one period before. 
However, we know that different time series have different periods, which means to have a good prediction, the model should be able to look back different numbers of time steps for different series. 
From this point, it is clear that the failure of traditional attention mechanisms comes from extracting only one previous time step while discounting the information in other time steps.
On the other hand, our attention mechanism attends on the features extracted from row vectors of RNN hidden states by CNN filters, which enables the model to select relevant information across multiple time steps.

The aforementioned explanation is verified by the left plot in Figure~\ref{fig:toy}, where we observe that the performance of the LSTM with Luong attention is poor when $D \gg 1$, compared to the others.
Notice that all models have similar amount of parameters, which implies that the LSTM without attention has a larger hidden size when compared to the LSTM with Luong attention.
Consequently, the LSTM without attention outperforms the LSTM with Luong attention when $D \gg 1$, because the larger hidden size helps the model to make prediction while the Luong attention is nearly useless.
On the contrary, our attention is useful, so the LSTM with our attention is better than the LSTM without attention on average, even though its hidden size is smaller.
Also, removing the CNN from our attention, which results in the same model as the ``Sigmoid - W/o CNN'' cell in Table~\ref{tab:ablation}, does not affect the performance, which implies that our feature-wise attention is indispensable.

\subsection{Influence of interdependencies}
When there are interdependencies in MTS data, it is desirable to leverage the interdependencies to further improve forecasting accuracy.
The right plot in Figure~\ref{fig:toy} shows that both the LSTM with Luong attention and the LSTM without attention do not benefit from the added interdependencies, since the loss values remain the same.
On the other hand, the loss of the LSTM with the proposed attention is lower when there are interdependencies, which suggests that our attention successfully utilized the interdependencies to facilitate MTS forecasting.
Again, removing the CNN from our attention does not affect the performance in this case.

\section{Experiments and Analysis}\label{sec:experiments_and_analysis}

In this section, we first describe the datasets upon which we conducted our experiments.
Next, we present our experimental results and a visualization of the prediction against LSTNet.
Then, we discuss the ablation study.
Finally, we analyze in what sense the CNN filters resemble the bases in DFT.

\subsection{Datasets}
To evaluate the effectiveness and generalization ability of the proposed attention mechanism, we used two dissimilar types of datasets: typical MTS datasets and polyphonic music datasets.

The typical MTS datasets are published by~\cite{LSTNet}; there are four datasets:
\begin{itemize}
    \item Solar Energy\footnote{\scriptsize \url{http://www.nrel.gov/grid/solar-power-data.html}}: the solar power production data from photovoltaic plants in Alabama State in 2006.
    \item Traffic\footnote{\scriptsize \url{http://pems.dot.ca.gov}}: two years (2015--2016) of data provided by the California Department of Transportation that describes the road occupancy rate (between 0 and 1) on San Francisco Bay area freeways.
    \item Electricity\footnote{\scriptsize \url{https://archive.ics.uci.edu/ml/datasets/ElectricityLoadDiagrams20112014}}: a record of the electricity consumption of 321 clients in kWh.
    \item Exchange Rate: the exchange rates of eight foreign countries (Australia, British, Canada, China, Japan, New Zealand, Singapore, and Switzerland) from 1990 to 2016.
\end{itemize}
These datasets are real-world data that contains both linear and non-linear interdependencies.
Moreover, the Solar Energy, Traffic, and Electricity datasets exhibit strong periodic patterns indicating daily or weekly human activities.
According to the authors of LSTNet, each time series in all datasets have been split into training ($60\%$), validation ($20\%$), and testing set ($20\%$) in chronological order.

In contrast, the polyphonic music datasets introduced below are much more complicated, in the sense that no apparent linearity or repetitive patterns exist:
\begin{itemize}
    \item MuseData~\cite{musedata}: a collection of musical pieces from various classical music composers in MIDI format. 
    \item LPD-5-Cleansed~\cite{musegan_0,musegan_1}: $21,425$ multi-track piano-rolls that contain drums, piano, guitar, bass, and strings.
\end{itemize}
To train models on these datasets, we consider each played note as 1 and 0 otherwise (i.e., a musical rest), and set one beat as one time step as shown in Table~\ref{tab:data-stats}.
Given the played notes of 4 bars consisting of 16 beats, the task is to predict whether each pitch at the next time step is played or not.
For training, validation, and testing sets, we follow the original MuseData separation, which is divided into 524 training pieces, 135 validation pieces, and 124 testing pieces.
LPD-5-Cleansed, however, was not split in previous work~\cite{musegan_0,musegan_1}; thus we randomly split it into training ($80\%$), validation ($10\%$), and testing ($10\%$) sets.
The size of LPD-5-Cleansed dataset is much larger than others, so we decided to use a smaller validation and testing set.

The main difference between typical MTS datasets and polyphonic music datasets is that scalars in typical MTS datasets are continuous but scalars in polyphonic music datasets are discrete (either 0 or 1).
The statistics of both the typical MTS datasets and polyphonic music datasets are summarized in Table~\ref{tab:data-stats}.

\begin{table}[t]
    \footnotesize
    \centering
    \begin{tabular}{|c||c|c|c|c|}
    \hline
    Dataset       &   $L$  & $D$ & $S$       & $B$ \\
    \hline 
    \hline 
    Solar Energy  & 52,560 & 137 & 10 minutes & 172 M\\
    \hline
    Traffic       & 17,544 & 862 & 1 hour     & 130 M \\
    \hline
    Electricity   & 26,304 & 321 & 1 hour     & 91 M \\
    \hline
    Exchange Rate &  7,588 &  8  & 1 day      & 534 K \\
    \hline 
    MuseData       & 216--102,552   & 128 & 1 beat & 4.9 M \\ 
    \hline 
    LPD-5-Cleansed & 1,072--1,917,952 & 128 & 1 beat & 1.7 G \\
    \hline
    \end{tabular}
    \caption{Statistics of all datasets, where $L$ is the length of the time series, $D$ is the number of time series, $S$ is the sampling spacing, and $B$ is size of the dataset in bytes. MuseData and LPD-5-Cleansed both have various-length time series since the length of music pieces varies.}
    \label{tab:data-stats}
\end{table}

\subsection{Methods for Comparison}
We compared the proposed model with the following methods on the typical MTS datasets:
\begin{itemize}
    \item AR: standard autoregression model.
    \item LRidge: VAR model with L2-regularization: the most popular model for MTS forecasting.
    \item LSVR: VAR model with SVR objective function~\cite{SVM}.
    \item GP: Gaussian process model~\cite{Gaussian-Process_1,Gaussian-Process_2}.
    \item SETAR: Self-exciting threshold autoregression model, a classical univariate non-linear model~\cite{SETAR}.
    \item LSTNet-Skip: LSTNet with recurrent-skip layer.
    \item LSTNet-Attn: LSTNet with attention layer.
\end{itemize}
AR, LRidge, LSVR, GP and SETAR are traditional baseline methods, whereas LSTNet-Skip and LSTNet-Attn are state-of-the-art methods based on deep neural networks.

However, as both traditional baseline methods and LSTNet are ill-suited to polyphonic music datasets due to their non-linearity and the lack of periodicity, we use LSTM and LSTM with Luong attention as the baseline models to evaluate the proposed model on polyphonic music datasets:
\begin{itemize}
    \item LSTM: RNN cells as introduced in Section~\ref{sec:preliminaries}.
    \item LSTM with Luong attention: LSTM with an attention mechanism scoring function of which $f(h_i, h_t) = (h_i)^\top W h_t$, where $W \in \mathbb{R}^{m \times m}$~\cite{luong}.
\end{itemize}

\subsection{Model Setup and Parameter Settings}
For all experiments, 
  we used LSTM units in our RNN models,  
and fixed the number of CNN filters at 32. 
Also, inspired by LSTNet, we included an autoregression component in our model when training and testing on typical MTS datasets.

For typical MTS datasets, we conducted a grid search over tunable parameters as done with LSTNet.
Specifically, on Solar Energy, Traffic, and Electricity, the range for window size $w$ was $\{24, 48, 96, 120, 144, 168\}$, the range for the number of hidden units $m$ was $\{25, 45, 70\}$, and the range for the step of the exponential learning rate decay with a rate of 0.995 was $\{200, 300, 500, 1000\}$.
On Exchange Rate, these three parameters were $\{30, 60\}$, $\{6, 12\}$, and $\{120, 200\}$, respectively.
Two types of data normalization were also viewed as part of the grid search: one normalized each time series by the maximum value in itself, and the other normalized every time series by the maximum value over the whole dataset.
Lastly, we used the absolute loss function and Adam with a $10^{-3}$ learning rate on Solar Energy, Traffic, and Electricity, and a $3 \cdot 10^{-3}$ learning rate on Exchange Rate.
For AR, LRidge, LSVR and GP, we followed the parameter settings as reported in the LSTNet paper~\cite{LSTNet}.
For SETAR, we searched the embedding dimension over \{24,48,96,120,144,168\} for Solar Energy, Traffic, and Electricity, and fixed the embedding dimension to 30 for Exchange Rate.
The two different setups between our method and LSTNet are (1)we have two data normalization methods to choose from, whereas LSTNet only used the first type of data normalization; and (2) the grid search over the window size $w$ is different.

For models used for the polyphonic music datasets, including the baselines and proposed models in the following subsections,
we used 3 layers for all RNNs, as done in~\cite{tonnetz}, and fixed the trainable parameters to around $5 \cdot 10^6$ by adjusting the number of LSTM units to fairly compare different models.
In addition, we used the Adam optimizer with a $10^{-5}$ learning rate and a cross entropy loss function.

\begin{table}
    \scriptsize
    \centering
    
    \begin{tabular}{|c||*{8}{P{0.77cm}|}}
    \hline
    RAE & \multicolumn{4}{c|}{Solar Energy} & \multicolumn{4}{c|}{Traffic} \\
    \hline
    Horizon & 3 & 6 & 12 & 24 & 3 & 6 & 12 & 24 \\
    \hline
    \hline
    AR & 0.1846 & 0.3242 & 0.5637 & 0.9221 & 0.4491 & 0.4610 & 0.4700 & 0.4696 \\ 
    \hline
    LRidge & 0.1227 & 0.2098 & 0.4070 & 0.6977 & 0.4965 & 0.5115 & 0.5198 & 0.4846 \\
    \hline
    LSVR & 0.1082 & 0.2451 & 0.4362 & 0.6180 & 0.4629 & 0.5483 & 0.7454 & 0.4761 \\
    \hline
    GP & 0.1419 & 0.2189 & 0.4095 & 0.7599 & 0.5148 & 0.5759 & 0.5316 & 0.4829 \\
    \hline
    SETAR & 0.1285 & 0.1962 & 0.2611 & \underline{0.3147} & 0.3226 & 0.3372 & \underline{0.3368} & \underline{0.3348} \\
    \hline
    LSTNet-Skip & 0.0985 & 0.1554 & \underline{0.2018} & 0.3551 & 0.3287 & 0.3627 & 0.3518 & 0.3852 \\
    \hline
    LSTNet-Attn & \textbf{0.0900} & \underline{0.1332} & 0.2202 & 0.4308 & \underline{0.3196} & \underline{0.3277} & 0.3557 & 0.3666 \\
    \hline
    Our model &
    \underline{0.0918} {\hspace*{-8pt} \fontsize{6}{6} \selectfont $\pm$~\underline{0.0005}} & \textbf{0.1296 {\hspace*{-8pt} \fontsize{6}{6} \selectfont $\pm$~0.0008}} & \textbf{0.1902 {\hspace*{-8pt} \fontsize{6}{6} \selectfont $\pm$~0.0021}} & \textbf{0.2727 {\hspace*{-8pt} \fontsize{6}{6} \selectfont $\pm$~0.0045}} & \textbf{0.2901 {\hspace*{-8pt} \fontsize{6}{6} \selectfont $\pm$~0.0095}} & \textbf{0.2999 {\hspace*{-8pt} \fontsize{6}{6} \selectfont $\pm$~0.0022}} & \textbf{0.3112 {\hspace*{-8pt} \fontsize{6}{6} \selectfont $\pm$~0.0015}} & \textbf{0.3118 {\hspace*{-8pt} \fontsize{6}{6} \selectfont $\pm$~0.0034}} \\
    \hline
    \end{tabular}
    
    \vspace{5pt}
    
    \begin{tabular}{|c||*{8}{P{0.77cm}|}}
    \hline
    RAE & \multicolumn{4}{c|}{Electricity} & \multicolumn{4}{c|}{Exchange Rate} \\
    \hline
    Horizon & 3 & 6 & 12 & 24 & 3 & 6 & 12 & 24 \\
    \hline
    \hline
    AR & 0.0579 & 0.0598 & 0.0603 & 0.0611 & 0.0181 & 0.0224 & 0.0291 & \underline{0.0378} \\ 
    \hline
    LRidge & 0.0900 & 0.0933 & 0.1268 & 0.0779 & 0.0144 & 0.0225 & 0.0358 & 0.0602 \\
    \hline
    LSVR & 0.0858 & 0.0816 & 0.0762 & 0.0690 & 0.0148 & 0.0231 & 0.0360 & 0.0576 \\
    \hline
    GP & 0.0907 & 0.1137 & 0.1043 & 0.0776 & 0.0230 & 0.0239 & 0.0355 & 0.0547 \\
    \hline
    SETAR & \underline{0.0475} & \underline{0.0524} & \underline{0.0545} & 0.0565 & \textbf{0.0136} & \underline{0.0199} & \underline{0.0288} & 0.0425 \\
    \hline
    LSTNet-Skip & 0.0509 & 0.0587 & 0.0598 & \underline{0.0561} & 0.0180 & 0.0226 & 0.0296 & \underline{0.0378} \\
    \hline
    LSTNet-Attn & 0.0515 & 0.0543 & 0.0561 & 0.0579 & 0.0229 & 0.0269 & 0.0384 & 0.0517 \\
    \hline
    Our model &
    \textbf{0.0463 {\hspace*{-8pt} \fontsize{6}{6} \selectfont $\pm$~0.0007}} & \textbf{0.0491 {\hspace*{-8pt} \fontsize{6}{6} \selectfont $\pm$~0.0007}} & \textbf{0.0541 {\hspace*{-8pt} \fontsize{6}{6} \selectfont $\pm$~0.0006}} & \textbf{0.0544 {\hspace*{-8pt} \fontsize{6}{6} \selectfont $\pm$~0.0007}} & \underline{0.0139} {\hspace*{-8pt} \fontsize{6}{6} \selectfont $\pm$~\underline{0.0001}} & \textbf{0.0192 {\hspace*{-8pt} \fontsize{6}{6} \selectfont $\pm$~0.0002}} & \textbf{0.0280 {\hspace*{-8pt} \fontsize{6}{6} \selectfont $\pm$~0.0006}} & \textbf{0.0372 {\hspace*{-8pt} \fontsize{6}{6} \selectfont $\pm$~0.0005}} \\
    \hline
    \end{tabular}
    
    \vspace{5pt}
    
    \begin{tabular}{|c||*{8}{P{0.77cm}|}}
    \hline
    RSE & \multicolumn{4}{c|}{Solar Energy} & \multicolumn{4}{c|}{Traffic} \\
    \hline
    Horizon & 3 & 6 & 12 & 24 & 3 & 6 & 12 & 24 \\
    \hline
    \hline
    AR & 0.2435 & 0.3790 & 0.5911 & 0.8699 & 0.5991 & 0.6218 & 0.6252 & 0.6293 \\
    \hline
    LRidge & 0.2019 & 0.2954 & 0.4832 & 0.7287 & 0.5833 & 0.5920 & 0.6148 & 0.6025 \\
    \hline
    LSVR & 0.2021 & 0.2999 & 0.4846 & 0.7300 & 0.5740 & 0.6580 & 0.7714 & 0.5909 \\
    \hline
    GP & 0.2259 & 0.3286 & 0.5200 & 0.7973 & 0.6082 & 0.6772 & 0.6406 & 0.5995 \\
    \hline
    SETAR & 0.2374 & 0.3381 & 0.4394 & 0.5271 & \underline{0.4611} & \underline{0.4805} & \underline{0.4846} & \underline{0.4898} \\
    \hline
    LSTNet-Skip & 0.1843 & 0.2559 & \underline{0.3254} & 0.4643 & 0.4777 & 0.4893 & 0.4950 & 0.4973 \\
    \hline
    LSTNet-Attn & \underline{0.1816} & \underline{0.2538} & 0.3466 & \underline{0.4403} & 0.4897 & 0.4973 & 0.5173 & 0.5300 \\
    \hline
    Our model & \textbf{0.1803 {\hspace*{-8pt} \fontsize{6}{6} \selectfont $\pm$~0.0008}} & \textbf{0.2347 {\hspace*{-8pt} \fontsize{6}{6} \selectfont $\pm$~0.0017}} & \textbf{0.3234 {\hspace*{-8pt} \fontsize{6}{6} \selectfont $\pm$~0.0044}} & \textbf{0.4389 {\hspace*{-8pt} \fontsize{6}{6} \selectfont $\pm$~0.0084}} & \textbf{0.4487 {\hspace*{-8pt} \fontsize{6}{6} \selectfont $\pm$~0.0180}} & \textbf{0.4658 {\hspace*{-8pt} \fontsize{6}{6} \selectfont $\pm$~0.0053}} & \textbf{0.4641 {\hspace*{-8pt} \fontsize{6}{6} \selectfont $\pm$~0.0034}} & \textbf{0.4765 {\hspace*{-8pt} \fontsize{6}{6} \selectfont $\pm$~0.0068}} \\
    \hline
    \end{tabular}
    
    \vspace{5pt}
    
    \begin{tabular}{|c||*{8}{P{0.77cm}|}}
    \hline
    RSE & \multicolumn{4}{c|}{Electricity} & \multicolumn{4}{c|}{Exchange Rate} \\
    \hline
    Horizon & 3 & 6 & 12 & 24 & 3 & 6 & 12 & 24 \\
    \hline
    \hline
    AR & 0.0995 & 0.1035 & 0.1050 & 0.1054 & 0.0228 & 0.0279 & \underline{0.0353} & \underline{0.0445} \\
    \hline
    LRidge & 0.1467 & 0.1419 & 0.2129 & 0.1280 & 0.0184 & 0.0274 & 0.0419 & 0.0675 \\
    \hline
    LSVR & 0.1523 & 0.1372 & 0.1333 & 0.1180 & 0.0189 & 0.0284 & 0.0425 & 0.0662 \\
    \hline
    GP & 0.1500 & 0.1907 & 0.1621 & 0.1273 & 0.0239 & 0.0272 & 0.0394 & 0.0580 \\
    \hline
    SETAR & 0.0901 & 0.1020 & 0.1048 & 0.1009 & \underline{0.0178} & \underline{0.0250} & 0.0352 & 0.0497 \\
    \hline
    LSTNet-Skip & \underline{0.0864} & \underline{0.0931} & 0.1007 & \underline{0.1007} & 0.0226 & 0.0280 & 0.0356 & 0.0449 \\
    \hline
    LSTNet-Attn & 0.0868 & 0.0953 & \underline{0.0984} & 0.1059 & 0.0276 & 0.0321 & 0.0448 & 0.0590 \\
    \hline
    Our model & \textbf{0.0823 {\hspace*{-8pt} \fontsize{6}{6} \selectfont $\pm$~0.0012}} & \textbf{0.0916 {\hspace*{-8pt} \fontsize{6}{6} \selectfont $\pm$~0.0018}} & \textbf{0.0964 {\hspace*{-8pt} \fontsize{6}{6} \selectfont $\pm$~0.0015}} & \textbf{0.1006 {\hspace*{-8pt} \fontsize{6}{6} \selectfont $\pm$~0.0015}} & \textbf{0.0174 {\hspace*{-8pt} \fontsize{6}{6} \selectfont $\pm$~0.0001}} & \textbf{0.0241 {\hspace*{-8pt} \fontsize{6}{6} \selectfont $\pm$~0.0004}} & \textbf{0.0341 {\hspace*{-8pt} \fontsize{6}{6} \selectfont $\pm$~0.0011}} & \textbf{0.0444 {\hspace*{-8pt} \fontsize{6}{6} \selectfont $\pm$~0.0006}} \\
    \hline
    \end{tabular}
    
    \vspace{5pt}
    
    \begin{tabular}{|c||*{8}{P{0.77cm}|}}
    \hline
    CORR & \multicolumn{4}{c|}{Solar Energy} & \multicolumn{4}{c|}{Traffic} \\
    \hline
    Horizon & 3 & 6 & 12 & 24 & 3 & 6 & 12 & 24 \\
    \hline
    \hline
    AR & 0.9710 & 0.9263 & 0.8107 & 0.5314 & 0.7752 & 0.7568 & 0.7544 & 0.7519 \\
    \hline
    LRidge & 0.9807 & 0.9568 & 0.8765 & 0.6803 & 0.8038 & 0.8051 & 0.7879 & 0.7862 \\
    \hline
    LSVR & 0.9807 & 0.9562 & 0.8764 & 0.6789 & 0.7993 & 0.7267 & 0.6711 & 0.7850 \\
    \hline
    GP & 0.9751 & 0.9448 & 0.8518 & 0.5971 & 0.7831 & 0.7406 & 0.7671 & 0.7909 \\
    \hline
    SETAR & 0.9744 & 0.9436 & 0.8974 & 0.8420 & 0.8641 & 0.8506 & 0.8465 & 0.8443 \\
    \hline
    LSTNet-Skip & 0.9843 & 0.9690 & \underline{0.9467} & 0.8870 & \underline{0.8721} & \underline{0.8690} & \underline{0.8614} & \underline{0.8588} \\
    \hline
    LSTNet-Attn & \underline{0.9848} & \underline{0.9696} & 0.9397 & \underline{0.8995} & 0.8704 & 0.8669 & 0.8540 & 0.8429 \\
    \hline
    Our model & \textbf{0.9850 {\hspace*{-8pt} \fontsize{6}{6} \selectfont $\pm$~0.0001}} & \textbf{0.9742 {\hspace*{-8pt} \fontsize{6}{6} \selectfont $\pm$~0.0003}} & \textbf{0.9487 {\hspace*{-8pt} \fontsize{6}{6} \selectfont $\pm$~0.0023}} & \textbf{0.9081 {\hspace*{-8pt} \fontsize{6}{6} \selectfont $\pm$~0.0151}} & \textbf{0.8812 {\hspace*{-8pt} \fontsize{6}{6} \selectfont $\pm$~0.0089}} & \textbf{0.8717 {\hspace*{-8pt} \fontsize{6}{6} \selectfont $\pm$~0.0034}} & \textbf{0.8717 {\hspace*{-8pt} \fontsize{6}{6} \selectfont $\pm$~0.0021}} & \textbf{0.8629 {\hspace*{-8pt} \fontsize{6}{6} \selectfont $\pm$~0.0027}} \\
    \hline
    \end{tabular}
    
    \vspace{5pt}
    
    \begin{tabular}{|c||*{8}{P{0.77cm}|}}
    \hline
    CORR & \multicolumn{4}{c|}{Electricity} & \multicolumn{4}{c|}{Exchange Rate} \\
    \hline
    Horizon & 3 & 6 & 12 & 24 & 3 & 6 & 12 & 24 \\
    \hline
    \hline
    AR & 0.8845 & 0.8632 & 0.8591 & 0.8595 & 0.9734 & 0.9656 & 0.9526 & 0.9357 \\
    \hline
    LRidge & 0.8890 & 0.8594 & 0.8003 & 0.8806 & \underline{0.9788} & \textbf{0.9722} & 0.9543 & 0.9305 \\
    \hline
    LSVR & 0.8888 & 0.8861 & 0.8961 & 0.8891 & 0.9782 & 0.9697 & \underline{0.9546} & \underline{0.9370} \\
    \hline
    GP & 0.8670 & 0.8334 & 0.8394 & 0.8818 & 0.8713 & 0.8193 & 0.8484 & 0.8278 \\
    \hline
    SETAR & \underline{0.9402} & \underline{0.9294} & \underline{0.9202} & \textbf{0.9171} & 0.9759 & 0.9675 & 0.9518 & 0.9314 \\
    \hline
    LSTNet-Skip & 0.9283 & 0.9135 & 0.9077 & 0.9119 & 0.9735 & 0.9658 & 0.9511 & 0.9354 \\
    \hline
    LSTNet-Attn & 0.9243 & 0.9095 & 0.9030 & 0.9025 & 0.9717 & 0.9656 & 0.9499 & 0.9339 \\
    \hline
    Our model & \textbf{0.9429 {\hspace*{-8pt} \fontsize{6}{6} \selectfont $\pm$~0.0004}} & \textbf{0.9337 {\hspace*{-8pt} \fontsize{6}{6} \selectfont $\pm$~0.0011}} & \textbf{0.9250 {\hspace*{-8pt} \fontsize{6}{6} \selectfont $\pm$~0.0013}} & \underline{0.9133} {\hspace*{-8pt} \fontsize{6}{6} \selectfont $\pm$~\underline{0.0008}} & \textbf{0.9790 {\hspace*{-8pt} \fontsize{6}{6} \selectfont $\pm$~0.0003}} & \underline{0.9709} {\hspace*{-8pt} \fontsize{6}{6} \selectfont $\pm$~\underline{0.0003}} & \textbf{0.9564 {\hspace*{-8pt} \fontsize{6}{6} \selectfont $\pm$~0.0005}} & \textbf{0.9381 {\hspace*{-8pt} \fontsize{6}{6} \selectfont $\pm$~0.0008}} \\
    \hline
    \end{tabular}
    \caption{Results on typical MTS datasets using RAE, RSE and CORR as metrics. Best performance in boldface; second best performance is underlined. We report the mean and standard deviation of our model in ten runs. All numbers besides the results of our model is referenced from the paper of LSTNet~\cite{LSTNet}.}
    \label{tab:time-series}
\end{table}

\subsection{Evaluation Metrics}

On typical MTS datasets, since we compared the proposed model with LSTNet, we followed the same evaluation metrics: RAE, RSE and CORR.
The first metric is the relative absolute error (RAE), which is defined as
{ 
\begin{equation}
\text{RAE} = \frac{\sum\limits_{t=t_0}^{t_1} \sum\limits_{i=1}^{n} |(y_{t,i} - \hat{y}_{t,i})|}{\sum\limits_{t=t_0}^{t_1} \sum\limits_{i=1}^n |\hat{y}_{t,i} - \overline{\hat{y}_{t_0:t_1,1:n}}|}.
\end{equation}
}
The next metric is the root relative squared error (RSE):
{ 
\begin{equation}
\text{RSE} = \frac{\sqrt{\sum\limits_{t=t_0}^{t_1} \sum\limits_{i=1}^{n} (y_{t,i} - \hat{y}_{t,i})^2}}{\sqrt{\sum\limits_{t=t_0}^{t_1} \sum\limits_{i=1}^n (\hat{y}_{t,i} - \overline{\hat{y}_{t_0:t_1,1:n}})^2}},
\end{equation}
}
and finally the third metric is the empirical correlation coefficient (CORR):
{ 
\begin{equation}
\text{CORR} = \frac{1}{n} \sum\limits_{i=1}^n \frac{\sum\limits_{t=t_0}^{t_1} (y_{t,i} - \overline{y_{t_0:t_1,i}})(\hat{y}_{t,i} - \overline{\hat{y}_{t_0:t_1,i}})}{\sqrt{\sum\limits_{t=t_0}^{t_1} (y_{t,i} - \overline{y_{t_0:t_1,i}})^2 \sum\limits_{t=t_0}^{t_1}(\hat{y}_{t,i} - \overline{\hat{y}_{t_0:t_1,i}})^2}},
\end{equation}
}
where $y, \hat{y}$ is defined in Section~\ref{subsec:problem_formulation}, $\hat{y}_t, \forall t \in [t_0, t_1]$ is the ground-truth value of the testing data, and $\overline{y}$ denotes the mean of set $y$.
RAE and RSE both disregards data scale and is a normalized version of the mean absolute error (MAE) and the root mean square error (RMSE), respectively.
For RAE and RSE, the lower the better, whereas for CORR, the higher the better.

To decide which model is better on polyphonic music datasets,
we use validation loss (negative log-likelihood), precision, recall, and F1 score as measurements
which are widely used in work on polyphonic music generation~\cite{musedata,tonnetz}.

\begin{figure}
 \centering
 \includegraphics[width=0.78\columnwidth]{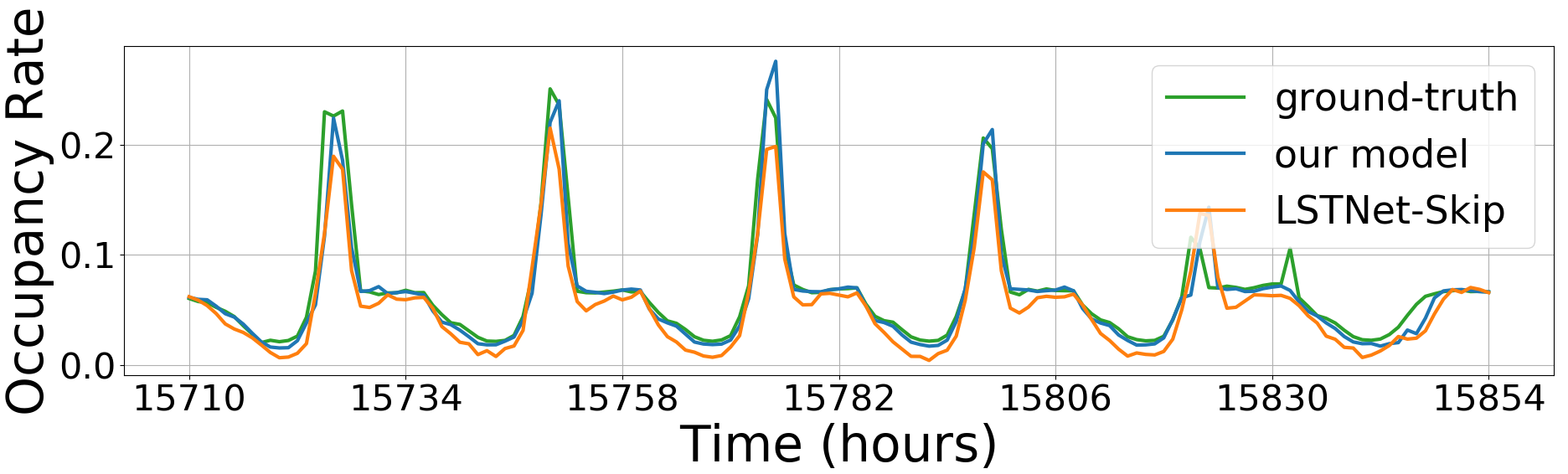}
 \caption{Prediction results for proposed model and LSTNet-Skip on Traffic testing set with 3-hour horizon. Proposed model clearly yields better forecasts around the flat line after the peak and in the valley.}
 \label{fig:visual_comp}
\end{figure}

\begin{figure}
  \centering
  \includegraphics[width=0.78\columnwidth]{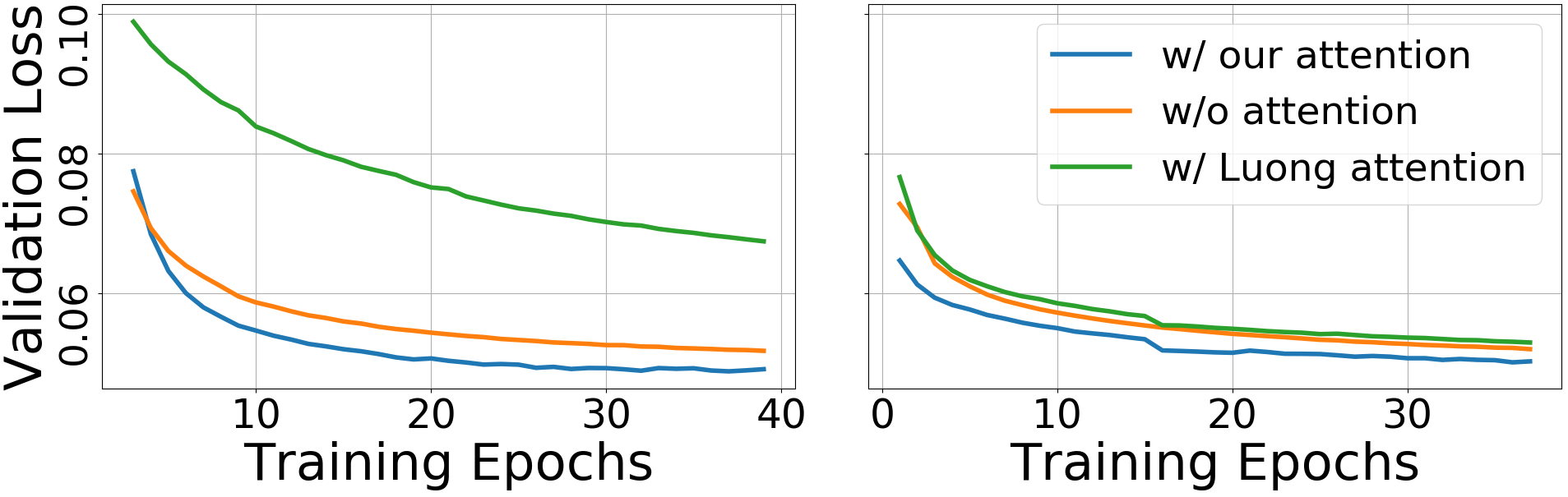}
  \caption{Validation loss under different training epochs on MuseData (left), and LPD-5-Cleansed (right).}
  \label{fig:learning_process}
\end{figure}

\subsection{Results on Typical MTS Datasets}
On typical MTS datasets, we chose the best model on the validation set using RAE/RSE/CORR as the metric for the testing set.
The numerical results are tabulated in Table~\ref{tab:time-series}, where the metric of the first two tables are RAE, followed by two tables of RSE metric, and ended by another two tables using CORR metric.
Both tables show that the proposed model outperforms almost all other methods on all datasets, horizons, and metrics.
Also, our models are able to deal with a wide range of dataset size, from the smallest 534 KB Exchange Rate dataset to the largest 172 MB Solar Energy dataset.
In these results, the proposed model consistently demonstrates its superiority for MTS forecasting.

In the comparison to LSTNet-Skip and LSTNet-Attn, the previous state-of-the-art methods, the proposed model exhibits superior performance, especially on Traffic and Electricity, which contain the largest amount of time series.
Moreover, on Exchange Rate, where no repetitive pattern exists, the proposed model is still the best overall; the performance of LSTNet-Skip and LSTNet-Attn fall behind traditional methods, including AR, LRidge, LSVR, GP, and SETAR.
In Figure~\ref{fig:visual_comp} we also visualize and compare the prediction of the proposed model and LSTNet-Skip.

In summary, the proposed model achieves state-of-the-art performance on both periodic and non-periodic MTS datasets.

\begin{table}
    \footnotesize
    \centering
    \begin{tabular}{|c||c|c|c|}
    \hline
                          & \multicolumn{3}{c|}{MuseData}                          \\
    \hline
    Metric                & Precision        & Recall           & F1               \\
    \hline
    \hline
    W/o attention         & 0.84009          & 0.67657          & 0.74952          \\
    \hline
    W/ Luong attention    & 0.75197          & 0.52839          & 0.62066          \\
    \hline
    W/ proposed attention & \textbf{0.85581} & \textbf{0.68889} & \textbf{0.76333} \\
    \hline
    \end{tabular}
    
    \vspace{5pt}
    
    \begin{tabular}{|c||c|c|c|}
    \hline
                          & \multicolumn{3}{c|}{LPD-5-Cleansed}                    \\
    \hline
    Metric                & Precision        & Recall           & F1               \\
    \hline
    \hline
    W/o attention         & 0.83794          & 0.73041          & 0.78049          \\
    \hline
    W/ Luong attention    & 0.83548          & 0.72380          & 0.77564          \\
    \hline
    W/ proposed attention & \textbf{0.83979} & \textbf{0.74517} & \textbf{0.78966} \\
    \hline
    \end{tabular}
    \caption{Precision, recall, and F1 score of different models on polyphonic music datasets.}
    \label{tab:music}
\end{table}

\subsection{Results on Polyphonic Music Datasets}
In this subsection, to further verify the efficacy and generalization ability of the proposed model to discrete data, we describe experiments conducted on polyphonic music datasets;
the results are shown in Figure~\ref{fig:learning_process} and Table~\ref{tab:music}.
We compared three RNN models: LSTM, LSTM with Luong attention, and LSTM with the proposed attention mechanism.
Figure~\ref{fig:learning_process} shows the validation loss across training epochs, and in Table~\ref{tab:music}, we use the models with the lowest validation loss to calculate precision, recall, and F1 score on the testing set.

From the results, we first verify our claim that the typical attention mechanism does not work on such tasks, as under similar hyperparameters and trainable weights, LSTM and the proposed model outperform such attention mechanisms.
In addition, the proposed model also learns more effectively compared to LSTM throughout the learning process and yields better performance in terms of precision, recall, and F1 score.

\begin{figure}[t]
  \centering
  \includegraphics[width=0.75\columnwidth]{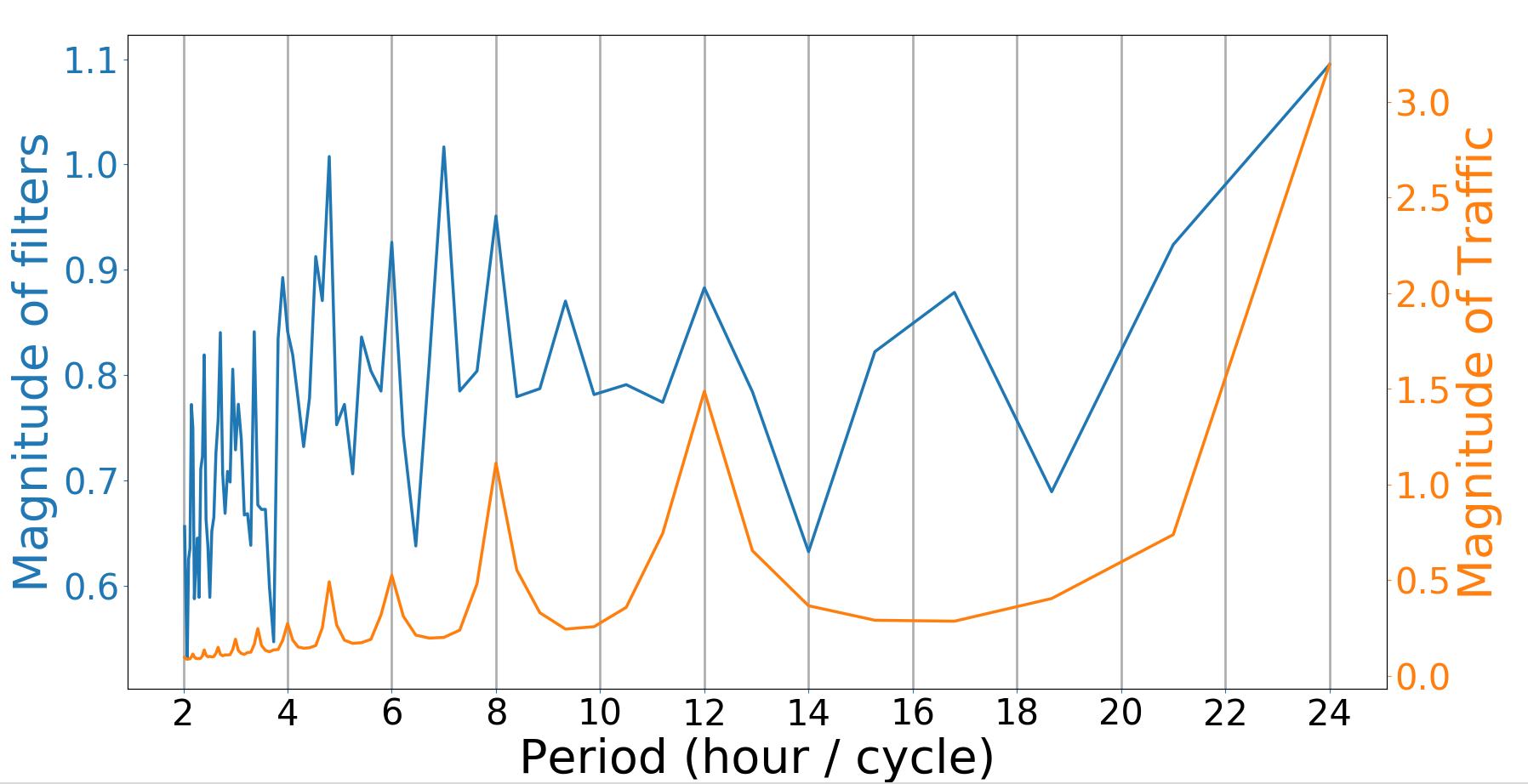}
  \caption{Magnitude comparison of (1) DFT of CNN filters trained on Traffic with a 3-hour horizon, and (2) every window of the Traffic dataset. To make the figure more intuitive, the unit of the horizontal axis is the period.}
  \label{fig:dft}
\end{figure}

\begin{figure}[t]
  \centering
  \includegraphics[width=0.72\columnwidth]{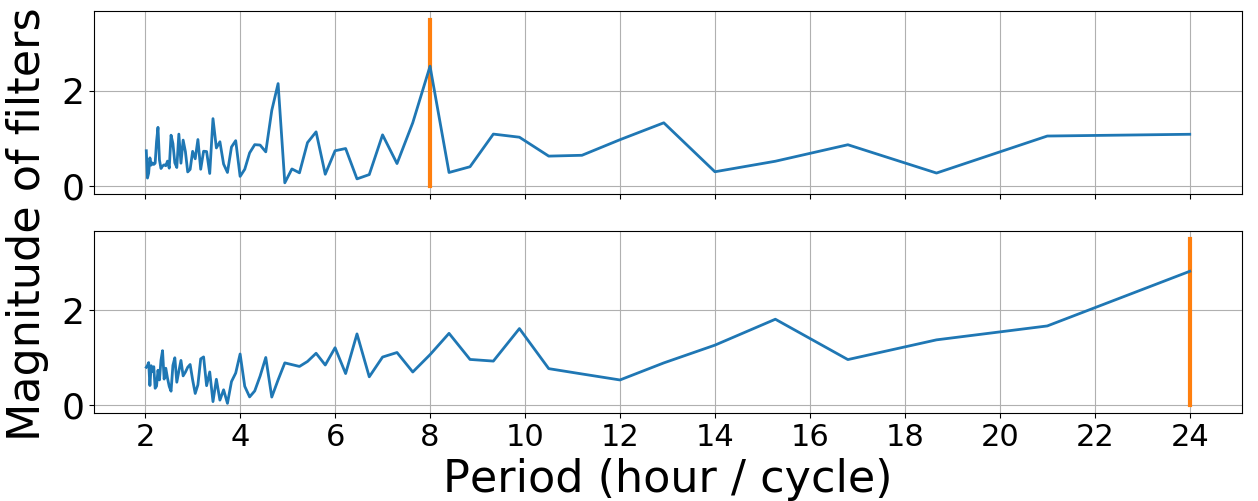}
  \caption{Two different CNN filters trained on Traffic with a 3-hour horizon, which detect different periods of temporal patterns.}
  \label{fig:four_filters}
\end{figure}

\subsection{Analysis of CNN Filters}
DFT is a variant of the Fourier transform (FT) which handles equally-spaced samples of a signal in time.
In the field of time series analysis, there is a wide body of work that utilizes FT or DFT to reveal important characteristics in time series~\cite{Fourier_0,Fourier_1}.
In our case, since the MTS data is also equally-spaced and discrete, we could apply DFT to analyze it.
However, in MTS data, there is more than one time series, so we naturally average the magnitude of the frequency components of every time series, and arrive at a single frequency domain representation.
We denote this the average discrete Fourier transform (avg-DFT).
The single frequency-domain representation reveals the prevailing frequency components of the MTS data.
For instance, it is reasonable to assume a notable 24-hour oscillation in Figure~\ref{fig:visual_comp}, which is verified by the avg-DFT of the Traffic dataset shown in Figure~\ref{fig:dft}.

Since we expect our CNN filters to learn temporal MTS patterns, the prevailing frequency components in the average CNN filters should be similar to that of the training MTS data.
Hence, we also apply avg-DFT on the $k = 32$ CNN filters that are trained on Traffic with a 3-hour horizon; in Figure~\ref{fig:dft} we plot the result alongside with the avg-DFT of every window of Traffic dataset.
Impressively, the two curves reach peaks at the same periods most of the time, which implies that the learned CNN filters resemble bases in DFT.
At the 24, 12, 8, and 6-hour periods, not only is the magnitude of the Traffic dataset at its peak, but the magnitudaaie of CNN filters also tops out. 
Moreover, in Figure~\ref{fig:four_filters}, we show that different CNN filters behave differently.
Some specialize at capturing long-term (24-hour) temporal patterns, while others are good at recognizing short-term (8-hour) temporal patterns.
As a whole, we suggest that the proposed CNN filters play the role of bases in DFT.
As demonstrated in the work by \cite{spectral_CNN}, such a ``frequency domain'' serves as a powerful representation for CNN to use in training and modeling.
Thus, LSTM relies on the frequency-domain information extracted by the proposed attention mechanism to accurately forecast the future.

\begin{table*}[t]
    \scriptsize
    \centering
    \begin{tabular}{|c||*{6}{P{1.3cm}|}}
    \hline
    Dataset & \multicolumn{3}{c|}{Solar Energy (Horizon = 24)} & \multicolumn{3}{c|}{Traffic (Horizon = 24)} \\
    \hline
            & Position         & Filter  & W/o CNN  & Position        & Filter  & W/o CNN \\
    \hline
    \hline
    Softmax & \underline{0.4397} {\hspace*{-8pt} \fontsize{6}{6} \selectfont $\pm$~\underline{0.0089}} & 0.4414 {\hspace*{-8pt} \fontsize{6}{6} \selectfont $\pm$~0.0093} & 0.4502 {\hspace*{-8pt} \fontsize{6}{6} \selectfont $\pm$~0.0099} & \textbf{0.4696 {\hspace*{-8pt} \fontsize{6}{6} \selectfont $\pm$~0.0062}} & 0.4832 {\hspace*{-8pt} \fontsize{6}{6} \selectfont $\pm$~0.0109} & 0.4810 {\hspace*{-8pt} \fontsize{6}{6} \selectfont $\pm$~0.0083} \\
    \hline
    Sigmoid & \textbf{0.4389 {\hspace*{-8pt} \fontsize{6}{6} \selectfont $\pm$~0.0084}} & 0.4598 {\hspace*{-8pt} \fontsize{6}{6} \selectfont $\pm$~0.0011} & 0.4639 {\hspace*{-8pt} \fontsize{6}{6} \selectfont $\pm$~0.0101} & \underline{0.4765} {\hspace*{-8pt} \fontsize{6}{6} \selectfont $\pm$~\underline{0.0068}} & 0.4785 {\hspace*{-8pt} \fontsize{6}{6} \selectfont $\pm$~0.0069} & 0.4803 {\hspace*{-8pt} \fontsize{6}{6} \selectfont $\pm$~0.0104} \\
    \hline
    Concat  & 0.4431 {\hspace*{-8pt} \fontsize{6}{6} \selectfont $\pm$~0.0100} & 0.4454 {\hspace*{-8pt} \fontsize{6}{6} \selectfont $\pm$~0.0093} & 0.4851 {\hspace*{-8pt} \fontsize{6}{6} \selectfont $\pm$~0.0049} & 0.4812 {\hspace*{-8pt} \fontsize{6}{6} \selectfont $\pm$~0.0082} & 0.4783 {\hspace*{-8pt} \fontsize{6}{6} \selectfont $\pm$~0.0077} & 0.4779 {\hspace*{-8pt} \fontsize{6}{6} \selectfont $\pm$~0.0073} \\
    \hline
    \end{tabular}
    
    \vspace{5pt}
    
    \begin{tabular}{|c||*{6}{P{1.3cm}|}}
    \hline
    Dataset & \multicolumn{3}{c|}{Electricity (Horizon = 24)} & \multicolumn{3}{c|}{MuseData}    \\
    \hline
            & Position         & Filter  & W/o CNN & Position         & Filter  & W/o CNN     \\
    \hline
    \hline
    Softmax & \textbf{0.0997 {\hspace*{-8pt} \fontsize{6}{6} \selectfont $\pm$~0.0012}} & 0.1007 {\hspace*{-8pt} \fontsize{6}{6} \selectfont $\pm$~0.0013} & 0.1010 {\hspace*{-8pt} \fontsize{6}{6} \selectfont $\pm$~0.0011} & \underline{0.04923} {\hspace*{-8pt} \fontsize{6}{6} \selectfont $\pm$~\underline{0.0037}} & 0.04929 {\hspace*{-8pt} \fontsize{6}{6} \selectfont $\pm$~0.0031} & 0.04951 {\hspace*{-8pt} \fontsize{6}{6} \selectfont $\pm$~0.0041} \\
    \hline
    Sigmoid & \underline{0.1006} {\hspace*{-8pt} \fontsize{6}{6} \selectfont $\pm$~\underline{0.0015}} & 0.1022 {\hspace*{-8pt} \fontsize{6}{6} \selectfont $\pm$~0.0009} & 0.1013 {\hspace*{-8pt} \fontsize{6}{6} \selectfont $\pm$~0.0011} & \textbf{0.04882 {\hspace*{-8pt} \fontsize{6}{6} \selectfont $\pm$~0.0031}} & 0.04958 {\hspace*{-8pt} \fontsize{6}{6} \selectfont $\pm$~0.0028} & 0.04979 {\hspace*{-8pt} \fontsize{6}{6} \selectfont $\pm$~0.0027} \\
    \hline
    Concat  & 0.1021 {\hspace*{-8pt} \fontsize{6}{6} \selectfont $\pm$~0.0017} & 0.1065 {\hspace*{-8pt} \fontsize{6}{6} \selectfont $\pm$~0.0029} & 0.1012 {\hspace*{-8pt} \fontsize{6}{6} \selectfont $\pm$~0.0008} & 0.05163 {\hspace*{-8pt} \fontsize{6}{6} \selectfont $\pm$~0.0040} & 0.05179 {\hspace*{-8pt} \fontsize{6}{6} \selectfont $\pm$~0.0036} & 0.05112 {\hspace*{-8pt} \fontsize{6}{6} \selectfont $\pm$~0.0027} \\
    \hline
    \end{tabular}
    \caption{Ablation Study. 
    Evaluation metric for Solar Energy, Traffic, and Electricity is RSE, and negative log-likelihood for MuseData. 
    We report the mean and standard deviation in ten runs.
    On each corpus, bold text represents the best and underlined text represents second best.}
    \label{tab:ablation}
\end{table*}

\subsection{Ablation Study}

In order to verify that the above improvement comes from each added component rather than a specific set of hyperparameters,
we conducted an ablation study on the Solar Energy, Traffic, Electricity, and MuseData datasets.
There were two main settings: one controlling how we attend to hidden states, $H$, of RNN and the other controlling how we integrate the scoring function $f$ into the proposed model, or even disable the function.
First, in the proposed method, we let the model attend to values of various filters on each position ($H^C_i$); we can also consider attending to values of the same filters at various positions ($(H^C)^\top_i$) or row vectors of $H$ ($H^\top_i$).
These three different approaches correspond to the column headers in Table~\ref{tab:ablation}: ``Position'', ``Filter'', and ``Without CNN''.
Second, whereas in the typical attention mechanism, softmax is usually used on the output value of scoring function $f$ to extract the most relevant information, we use sigmoid as our activation function.
Therefore, we compare these two different functions.
Another possible structure for forecasting is to concatenate all previous hidden states and let the model automatically learn which values are important.
Taking these two groups of settings into consideration, we trained models with all combinations of possible structures on these four datasets.

The MuseData results show that the model with sigmoid activation and attention on $H^C_i$ (position) is clearly the best, which suggests that the proposed model is reasonably effective for forecasting.
No matter which proposed component is removed from the model, performance drops.
For example, using softmax instead of sigmoid raises the negative log-likelihood from $0.04882$ to $0.04923$; we obtain a even worse model with a negative log-likelihood of $0.4979$ if we do not use CNN filters.
In addition, we note no significant improvement between the proposed model and that model using softmax on the first three datasets in Table~\ref{tab:ablation}: Solar Energy, Traffic, and Electricity.
This is not surprising, given our motivation for using sigmoid, as explained in Section~\ref{ssec:proposed_attention}.
Originally, we expected CNN filters to find basic patterns and expected the sigmoid function to help the model to combine these patterns into one that helps.
However, due to the strongly periodic nature of these three datasets, it is possible that using a small number of basic patterns is sufficient for good prediction.
Overall, however, the proposed model is more general and yields stable and competitive results across different datasets.

\section{Conclusions}\label{sec:conclusions}
In this paper, we focus on MTS forecasting and propose a novel temporal pattern attention mechanism which removes the limitation of typical attention mechanisms on such tasks.
We allow the attention dimension to be feature-wise in order for the model learn interdependencies among multiple variables not only within the same time step but also across all previous times and series.
Our experiments on both toy examples and real-world datasets strongly support this idea and show that the proposed model achieves state-of-the-art results.
In addition, the visualization of filters also verifies our motivation in a more understandable way to human beings.


\bibliographystyle{spbasic}      
\bibliography{template}   


\end{document}